\documentclass[10pt,twocolumn,letterpaper]{article}
\usepackage[ruled,vlined]{algorithm2e}
\usepackage{amsmath}
\usepackage{cvpr}              

\usepackage[accsupp]{axessibility}

\usepackage{pifont}

\usepackage{multirow}

\definecolor{cvprblue}{rgb}{0.21,0.49,0.74}
\usepackage[pagebackref,breaklinks,colorlinks,allcolors=cvprblue]{hyperref}
\usepackage{xcolor} 
\usepackage{colortbl}     
\definecolor{cGrey}{HTML}{eff3fa}
\newcommand{\model}{\textit{SAIL}}

\definecolor{mjjeonc}{RGB}{34,139,34}

\AtEndPreamble{
    \crefname{section}{Sec.}{Secs.}
    \Crefname{section}{Section}{Sections}
    \Crefname{table}{Table}{Tables}
    \crefname{table}{Tab.}{Tabs.}
    \Crefname{figure}{Figure}{Figures}
}

\def\paperID{5504} 
\def\confName{CVPR}
\def\confYear{2026}

\title{SAIL: Similarity-Aware Guidance and Inter-Caption Augmentation-based Learning for Weakly-Supervised Dense Video Captioning}

\author{Ye-Chan Kim \quad SeungJu Cha \quad Si-Woo Kim \quad Minju Jeon \quad Hyungee Kim \quad Dong-Jin Kim\\
  Hanyang University, South Korea \\ 
    \tt\small{\{dpcksdl78, sju9020, boreng0817, mnju5026, khjiiii2002, djdkim\}}@hanyang.ac.kr 
    }

\begin{document}
\maketitle
 \begin{abstract}
Weakly-Supervised Dense Video Captioning aims to localize and describe events in videos trained only on caption annotations, without temporal boundaries. Prior work introduced an implicit supervision paradigm based on Gaussian masking and complementary captioning. 
However, existing method focuses merely on generating non-overlapping masks without considering their semantic relationship to corresponding events, resulting in simplistic, uniformly distributed masks that fail to capture semantically meaningful regions. 
Moreover, relying solely on ground-truth captions leads to sub-optimal performance due to the inherent sparsity of existing datasets.
In this work, we propose $\model$, which constructs semantically-aware masks through cross-modal alignment. 
Our similarity-aware training objective guides masks to emphasize video regions with high similarity to their corresponding event captions.
Furthermore, to guide more accurate mask generation under sparse annotation settings, we introduce an LLM-based augmentation strategy that generates synthetic captions to provide additional alignment signals.
These synthetic captions are incorporated through an inter-mask mechanism, providing auxiliary guidance for precise temporal localization without degrading the main objective.
Experiments on ActivityNet Captions and YouCook2 demonstrate state-of-the-art performance on both captioning and localization metrics.
\end{abstract}    
 \section{Introduction}
\label{sec:intro}
\begin{figure}[t]
\begin{center}
\includegraphics[width=1\columnwidth]{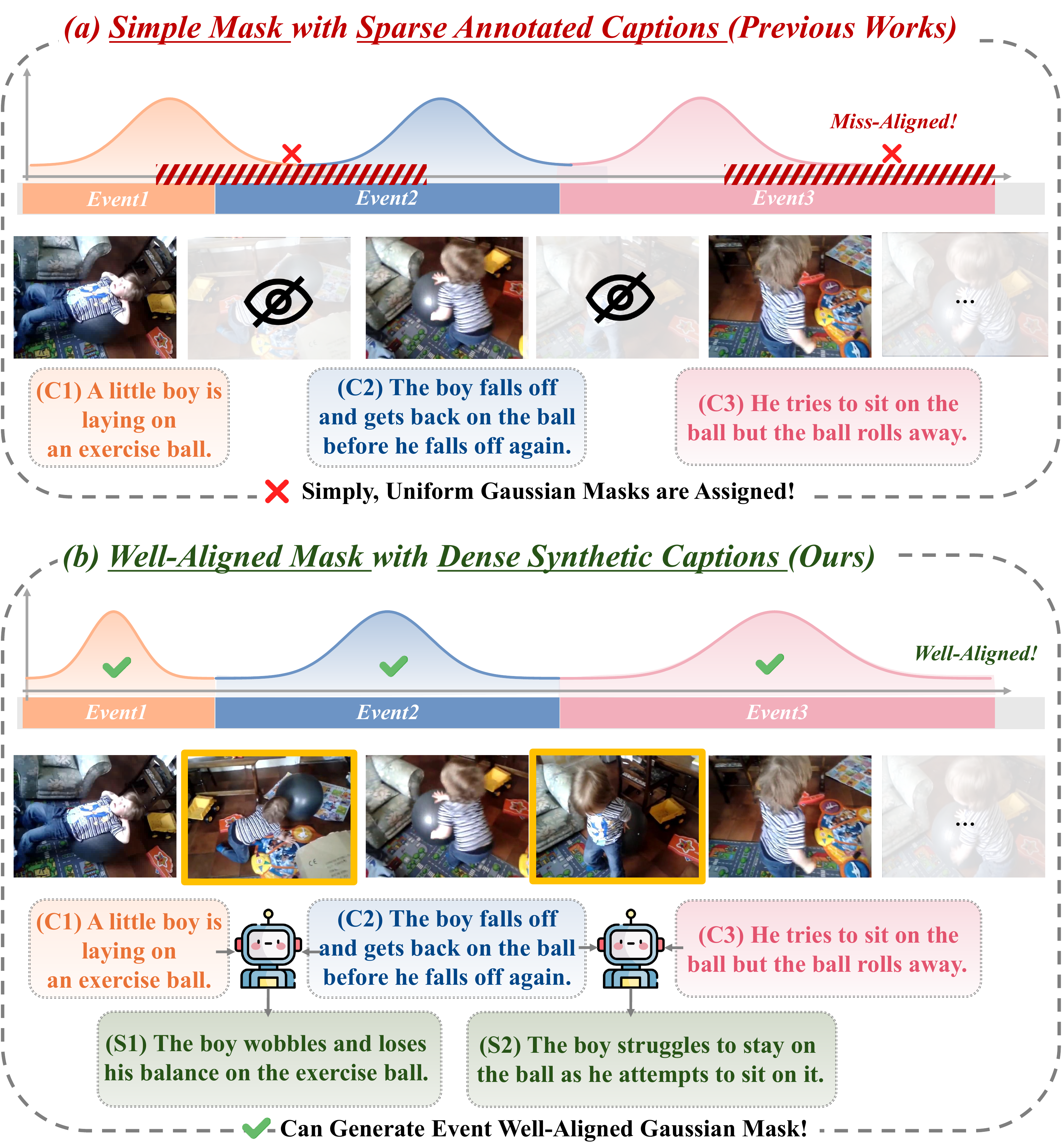}
\end{center}
\caption{(a) Previous work generates masks that simply cover different temporal regions without considering semantic alignment with corresponding events. (b) Our proposed method leverages cross-modal similarity to guide masks toward event-relevant regions and addresses caption sparsity through LLM augmentation.}
\label{fig:teaser}
\end{figure}
In diverse multimodal applications~\cite{scalediff,verbdiff,vipcap}, Dense Video Captioning (DVC) is one of the key research area in the video domain.
Unlike conventional video captioning~\cite{gao2017svc, chen2017svc, wang2018svc}, DVC requires jointly localizing events and generating captions in long, untrimmed videos.
Although diverse DVC methods have been proposed, research has mainly focused on fully-supervised settings~\cite{vid2seq,jeon2025sali4vid,kim2025hicm2,cm2}. 
These methods depend on exhaustively labeled data with precise event boundaries and descriptive captions for each event, making annotation prohibitively expensive and labor-intensive for real-world applications.
To address these issues, several Weakly-Supervised Dense Video Captioning (WSDVC) studies have emerged~\cite{WSDEC,pws-dvc,ge2025implicit,ECG}, which do not rely on event boundaries and only use captions for training.

\begin{figure*}[t]
\includegraphics[width=\textwidth]{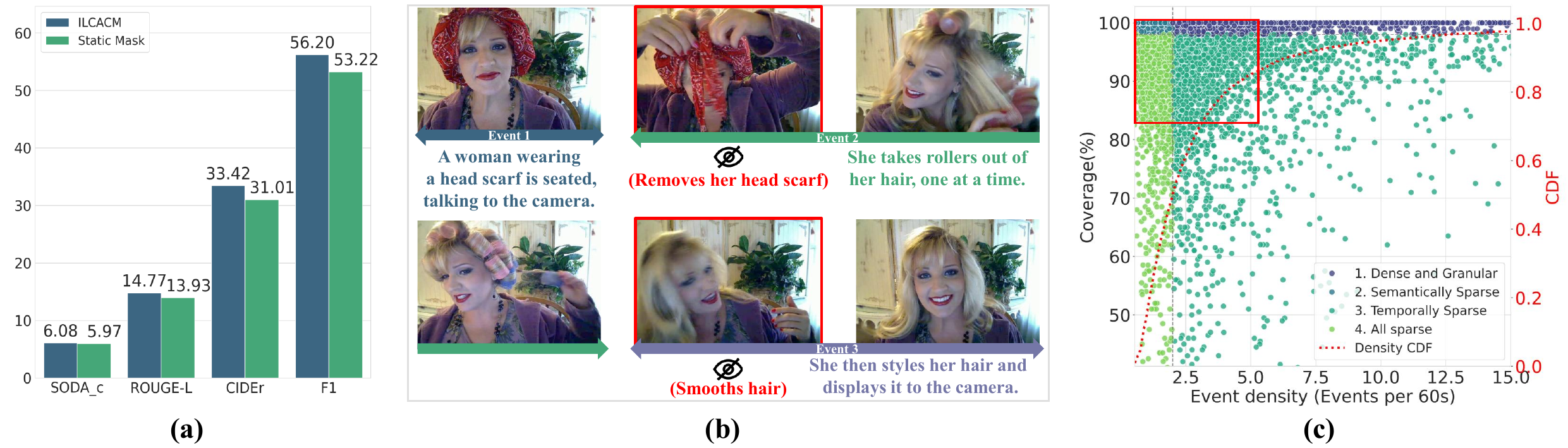}
\caption{(a) The fixed mask baseline applies Gaussian masks with equal width uniformly distributed according to the number of events. It performs comparably to the existing method~\cite{ge2025implicit}.
(b) shows the dataset's annotation sparsity example, where potential events are missed.
(c) The majority of video samples have events spanning the entire duration but contain only a small number of events (red box).
}
\label{fig:problem_experiment}
\end{figure*}

Recently, ILCACM~\cite{ge2025implicit}, the current state-of-the-art WSDVC method, introduces a Gaussian masking strategy that applies differentiable masks to video features, enabling implicit event localization through complementary captioning.
Although it shows promising results, we observe that this method overlooks the semantic alignment between masked features and their corresponding events, which we believe is a critical aspect for learning discriminative event representations.
As shown in \Cref{fig:teaser}, when constructing masks for target events, the existing approach merely focuses on generating mutually exclusive masks without ensuring semantic consistency with corresponding events, often resulting in simplistic and uniformly distributed masks.
These \textit{simplistic and uniformly} distributed masks lead the model to fail to learn meaningful event emphasis, resulting in low-quality captions and inaccurate localization.
{As shown in \Cref{fig:problem_experiment} \textcolor{cvprblue}{(a)}, we find that even a fixed, non-trainable mask baseline performs comparably to the existing method~\cite{ge2025implicit} on both captioning and localization tasks.}
This indicates that the existing approach merely learns to emphasize distinct temporal regions without capturing semantic event relevance.

To address this issue, our primary goal is to construct genuinely informative masks for target events.
We define an informative mask as one that, rather than simply covering a distinct temporal region, highlights video segments whose features exhibit high cross-modal similarity with the corresponding event caption.
To achieve this, we leverage the strong cross-modal alignment of Vision-Language model CLIP~\cite{clip} and guide the model to generate masks based on cross-modal similarity signals rather than enforcing only non-overlap constraints.
We train the model to align masked video features with corresponding ground-truth captions while distinguishing non-corresponding ones. 
Despite the strong overall performance of the cross-modal similarity guide, we observe some failure cases when videos have very sparse annotations (e.g., only 1-2 events).
We hypothesize that these failures stem from the sparse characteristic of ground-truth annotations~\cite{AAAI2024exploitingauxcaption}, which leads to insufficient signal for the model to effectively learn the alignment between masked features and their corresponding captions.
For example, as illustrated in \Cref{fig:problem_experiment}\textcolor{cvprblue}{(b)}, a sample video from ActivityNet~\cite{ActivityNet} spans 235 seconds yet contains only three event annotations, leaving numerous potential events unannotated.
Also, as shown in~\Cref{fig:problem_experiment}\textcolor{cvprblue}{(c)}, although the majority of videos have annotations covering their entire duration, the event density remains consistently low, demonstrating that most videos show sparse event distributions.
In such sparse annotation regimes, we conjecture that over-reliance on only ground-truth caption guidance provides insufficient supervision for learning accurate mask-event alignment, especially in WSDVC.

To overcome these limitations, we introduce a Large-Language Model (LLM) based caption augmentation strategy that leverages recent advances in LLMs' contextual understanding and reasoning capabilities~\cite{gpt4,CoTLLM}. 
Specifically, we construct prompts using consecutive pairs of ground-truth captions and task the LLM with generating plausible descriptions for the temporal segments connecting them.
These synthetic captions are then interleaved with the original annotations, creating a dense supervision signal that captures a more diverse range of potential events.
By training on both real and synthetic captions, our cross-modal objective learns more nuanced similarity guidance, enabling masks to capture finer-grained event details.
To this end, we propose $\model$, \textbf{S}imilarity-\textbf{A}ware Guidance and \textbf{I}nter-Caption Augmentation-based \textbf{L}earning for WSDVC.
Our contributions are as follows:
\begin{itemize}
    \item We propose similarity-aware mask guide, which leverages the cross-modal alignment information to generate event-focused masks.
    \item We propose a novel method that generates synthetic captions via LLM and utilizes them as auxiliary supervision to address annotation sparsity, enabling the model to learn from dense alignment signals.
    \item We validate our method on ActivityNet and
    YouCook2, achieving state-of-the-art results in both
    the localization and captioning tasks.
\end{itemize}

 \section{Related Work}
\label{sec:related}
\subsection{Dense Video Captioning}
DVC is a challenging multi-task that requires the simultaneous execution of event localization and event captioning in untrimmed videos. In the early stages of DVC~\cite{ActivityNet,Wang_2018_CVPR,iashin2020multi}, many studies adopted a “localize-then-caption” strategy, where events are first localized and then a caption is generated for each event. However, this 2-stage approach overlooked the interaction between the localization and captioning tasks; also, they rely on complex, hand-crafted components and extensive hyper-parameter tuning.
To overcome these limitations, DETR-based transformer architecture~\cite{cm2,kim2025hicm2,jeon2025sali4vid,vid2seq} emerged to decode both event timestamps and captions in parallel.
\citet{vid2seq} proposed a method of pre-training on a large-scale narrated video dataset using ASR and then fine-tuning.
Based on the above pre-trained model,~\cite{cm2,kim2025hicm2,jeon2025sali4vid,vid2seq} enhances captioning ability by utilizing an external memory of captions.
Despite extensive research into various DVC methods, most studies still operate in fully supervised settings. 
This is a notable limitation, given the extremely high annotation cost associated with localizing and captioning events in video data.
\subsection{Weakly-Supervised Dense Video Captioning}
WSDVC aims to perform the DVC task without this temporal supervision.~\citet{WSDEC} first addressed WSDVC by decomposing the problem into sentence localization and event captioning.
Building upon this, subsequent research has largely revolved around cycle training systems~\cite{chidume1987iterative}. 
Among these,~\citet{EC-SL} introduced a concept learner approach and ~\citet{pws-dvc} further enhanced performance by leveraging the MSR-VTT dataset~\cite{xu2016msr} for pre-training.
Recently,~\citet{ge2025implicit} has proposed a method that implicitly performs localization and captioning through complementary masked captioning.
Although this complementary masked captioning has proven effective, its masking strategy is simplistic, as it merely ensures that the masks cover different regions.
To address this issue, we propose a method to ensure that the masked region is well-aligned with its corresponding caption.

 \begin{figure*}[t]
    \centering
\includegraphics[width=0.93\textwidth]{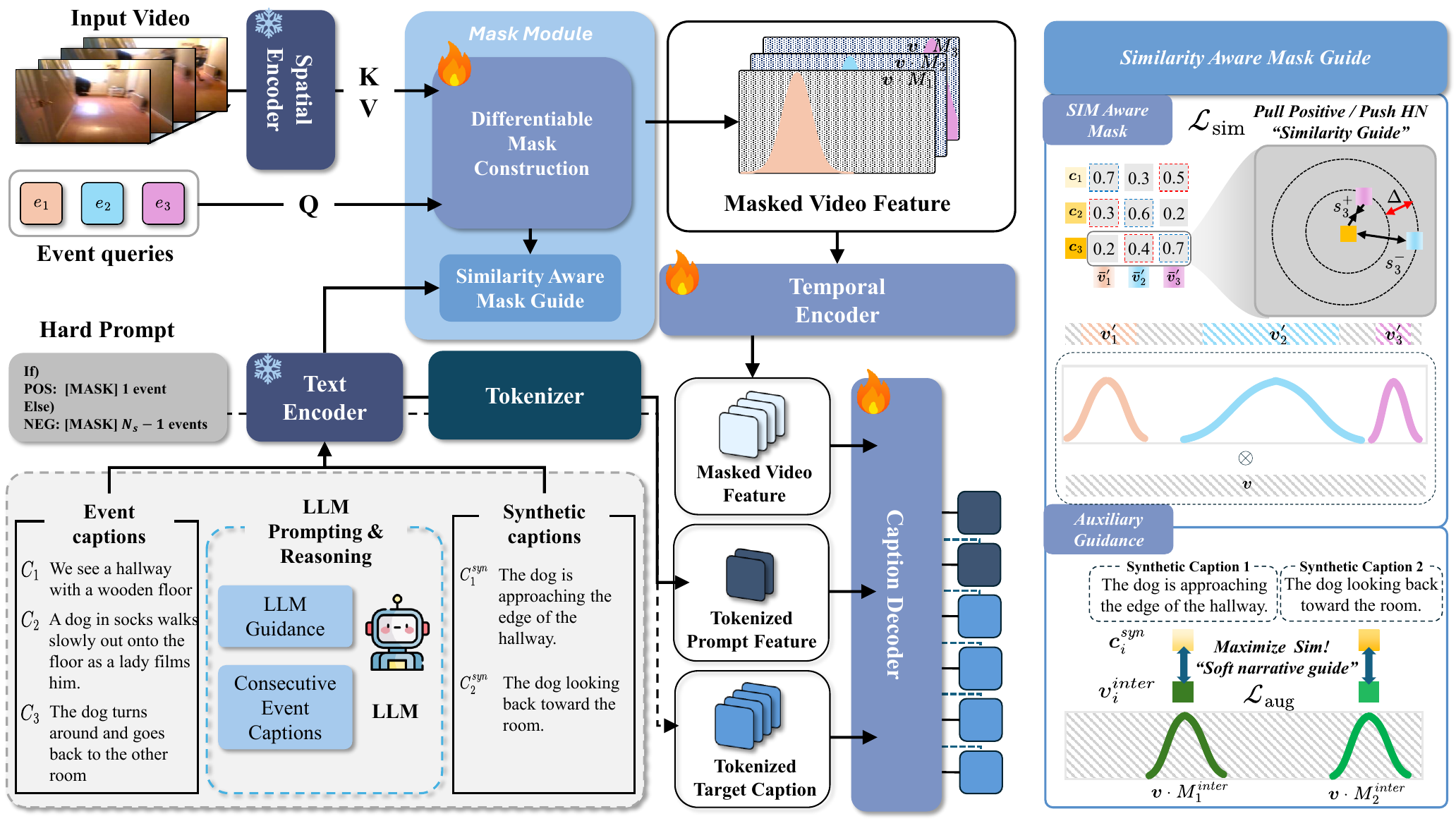}
    \caption{$\model$ Pipeline. Our method exploits cross-modal similarity to guide mask optimization toward increased alignment with event captions and further enriches supervision through LLM-generated synthetic captions.
    }
    \label{fig:main}
\end{figure*}
\label{sec:method}
\section{Method}

{In WSDVC, given a video $V = \{v_i\}_{i=1}^{N_v}$ with $N_v$ frames that encompass $N_s$ distinct events, the goal is to generate corresponding captions $\{C_i\}_{i=1}^{N_s}$ for each event and localize them temporally.}
As WSDVC is trained only with caption annotations, the model learns to infer temporal event boundaries by aligning video frames with their corresponding textual descriptions~\cite{ge2025implicit,WSDEC,pws-dvc}.

Our goal is to construct informative masks that emphasize the visual features most relevant to their corresponding event captions.
To achieve this, we simultaneously enforce semantic alignment via a cross-modal objective and improve mask quality by supplementing supervision signals with synthetic captions, as illustrated in ~\Cref{fig:main}. 

\subsection{Preliminaries} 
\noindent\textbf{Baseline.}
ILCACM~\cite{ge2025implicit} introduces a Gaussian masking strategy that applies differentiable masks to video features, enabling implicit event localization through complementary captioning. 
Specifically, for each learnable event query, a transformer-based mask decoder outputs an event-specific feature embedding. 
This embedding is then passed through two separate fully-connected layers to predict two scalar values for the $i$-th event: its temporal center $(c_i)$ and its width $(w_i)$. 
These values are then used to construct a Gaussian mask $M_i$ across the video's temporal axis $t$:
\begin{equation}
M_{i}(t) = \exp \left( - \frac{(t - c_{i})^{2}}{2(w_{i} / \tau)^{2}} \right),
\label{eq:mask}
\end{equation}
where $\tau$ is a temperature parameter that controls the sharpness of the Gaussian curve.
To ensure that masks cover different regions, ILCACM employs a mask diversity loss based on cosine similarity between mask distributions.

Utilizing these Gaussian masks, the model generates two complementary features: the ``positive" masked video features $(\boldsymbol{v} \cdot M_i)$ to produce a caption describing only the target event, while the ``negative" masked video features $(\boldsymbol{v} \cdot (1-M_i))$ are used to produce a caption describing all remaining events in the video,
where $\boldsymbol{v}$ are frame-level embeddings from the spatial encoder and then these masked features are fed into the video encoder.
The decoder takes as input these encoded visual features, along with prompts and target captions for both the positive and negative sets, and generates the corresponding captions in an auto-regressive manner.
The resulting outputs are used to compute the positive captioning loss $\mathcal{L}_{\text{pos}}$ and negative captioning loss $\mathcal{L}_{\text{neg}}$. 
By constraining these two captions to form the complete video description (i.e., $C=C_i \cup \{C_j\}_{j=1, j \neq i}^{N_s}$), the mask prediction module learns to localize events implicitly. 

\subsection{Similarity-Aware Mask Guide}
\label{sec:saliency}
As established in the introduction, our goal is to generate event masks that are not merely temporally separated but are semantically aligned with their corresponding event captions. 
To achieve this, we design a mask generation process that is directly guided by cross-modal similarity.

\noindent\textbf{Similarity-Aware Mask Training.}
We make the differentiable masks informative by designing a training objective that encourages them to highlight semantically salient regions. 
After a mask $M_i$ is generated, it is multiplied element-wise with the video features $\boldsymbol{v}$ to produce positive masked video features $\boldsymbol{v}'_i=(\boldsymbol{v}\cdot M_i)$. 
These masked features serve as the visual representation of the emphasized region for event $i$. 
To ensure $\boldsymbol{v}'_i$ truly captures the correct event content, we leverage event captions as semantic guidance. 
We maximize the cross-modal cosine similarity between the average pooled masked features $\bar{\boldsymbol{v}}_i'$ and their corresponding caption features  $\boldsymbol{c}_i$ while minimizing similarity to other event captions in the same video through a margin ranking loss~\cite{margin}: 
\begin{equation}
\mathcal{L}_{\text{sim}} = \frac{1}{B} \sum_{b=1}^{B} \frac{1}{N_s} \sum_{i=1}^{N_s} \max\left(0, \Delta - s^+_{b,i} + s^-_{b,i}\right),
\label{eq:saliency}
\end{equation}
where $s^+_{b,i} = \text{sim}(\bar{\boldsymbol{v}}'_{b,i}, \boldsymbol{c}_{b,i})$ is the positive pair similarity, $s^-_{b,i} = \max_{j \neq i} \text{sim}(\bar{\boldsymbol{v}}'_{b,i}, \boldsymbol{c}_{b,j})$ is the hard negative similarity within video $v$, $B$ is the batch size, $N_s$ is the number of events in video $v$, and $\Delta$ is a margin hyperparameter.
By optimizing $\mathcal{L}_{\text{sim}}$, each mask learns to emphasize video regions that are semantically closest to its corresponding caption, effectively aligning the masks with event-specific visual content.
For negative captioning, we follow the same approach as~\cite{ge2025implicit}, constructing inverse masks and training them in the same manner as described above.

\subsection{LLM-Based Caption Augmentation}
While $\mathcal{L}_{\text{sim}}$ enables the masks to localize semantically salient event regions, we observe that the model often generates coarse event boundaries, especially when captions are temporally sparse in a given video.
Empirically, we find that although event annotations may cover most of the video duration, they often consist of only a few coarse-grained descriptions with large temporal gaps between consecutive annotations (\Cref{fig:problem_experiment} \textcolor{cvprblue}{(c)}).
We hypothesize that such sparsity in temporal descriptions limits the model's ability to learn fine-grained event localization.

To better capture fine-grained event boundaries, we propose an LLM-based augmentation strategy that generates synthetic captions for intermediate events between consecutive ground-truth annotations.
{By leveraging the LLM's world knowledge and contextual reasoning~\cite{zeroreasoner,LMworld,CoTLLM}, we guide the model to infer plausible transitional events that bridge these temporal gaps, thereby providing denser and more informative supervision across the video timeline.}

\noindent\textbf{Transition Caption Generation.}
First, we design a carefully structured prompt to guide the LLM in generating high-quality synthetic captions.
Inspired by the effectiveness of role-based prompting in task-specific reasoning~\cite{CoTLLM}, we define the LLM's role as a ``Video Context Inference Expert" specializing in temporal event analysis.
The system prompt explicitly instructs the model to: (1) analyze both preceding and succeeding captions to understand narrative flow, (2) infer the most probable \textit{transitional action} or \textit{change of state} rather than merely rephrasing existing captions, and (3) maintain consistency in style, tone, and detail level with the original annotations.
This structured format helps the LLM focus on the specific temporal gap while avoiding extraneous information.
Then, we feed consecutive captions $C_i$ and $C_{i+1}$ along with the structured query into the LLM to generate a synthetic description $C^{syn}_i$ representing the transitional event.
We apply this generation process to each pair of consecutive captions in a video, producing $N_s - 1$ synthetic captions per video.
Further details about prompts are provided in the supplementary material.

\noindent\textbf{Auxiliary Guidance via Synthetic Captions.} 
The synthetic captions generated by LLM can provide a much denser, more fine-grained narrative guide. 
Although the synthetic captions contain plausible information, using them directly in the main contrastive loss $\mathcal{L}_{\text{sim}}$ (e.g., as hard negatives) could introduce noise and degrade performance (\Cref{table:synth_cap_utilize}).
Therefore, we propose to use these synthetic captions as a separate, auxiliary training signal. 
The goal is to guide the model to find visual evidence from these potential transitional events without strictly enforcing them. 
To achieve this, we first create a new set of “inter-masks" $(M^{inter}_i)$, designed to represent the temporal segments between the predicted ground-truth events masks $(M_i, M_{i+1})$. 
For each adjacent pair of predicted event centers $(c_i, c_{i+1})$, we define a new inter-mask center $c^{inter}_i$ as their average: $c^{inter}_i = \frac{c_i + c_{i+1}}{2}$.
The width for these inter-masks is set to a fixed hyperparameter $w^{inter}$. 
These parameters $(c^{inter}_i, w^{inter})$ are fed into the same Gaussian function~\Cref{eq:mask} to generate a total of $N_s-1$ new masks, 
$M^{inter}_i$.
We then apply these inter-masks to the video features to get the augmented masked features $\boldsymbol{v}'^{inter}_i=(\boldsymbol{v} \cdot M^{inter}_i)$, which are aggregated into representations. 
We apply average pooling to the augmented masked features, and $\mathcal{L}_{\text{aug}}$ encourages the resulting representations to align with their corresponding transition caption embeddings $\boldsymbol{c}^{syn}_i$:
\begin{equation}
\mathcal{L}_{\text{aug}} = \frac{1}{B} \sum_{b=1}^{B}\frac{1}{N_s-1} \sum_{i=1}^{N_s-1}\left( 1 - \text{sim}(\bar{\boldsymbol{v}}'^{inter}_{b,i}, \boldsymbol{c}^{syn}_{b,i}) \right).
\end{equation}
This loss acts as a soft narrative guide, encouraging the model to capture the transitional events. 
This in turn, helps the main predictors for $c_i$ and $c_{i+1}$ to be more precise, allowing the model to represent more fine-grained event boundaries.

\noindent\textbf{Final Objective.}
Our final training objective is a positive \& negative captioning  loss,  cross-modal similarity loss, and the auxiliary augmentation loss:
\begin{equation}
\mathcal{L} = \mathcal{L}_{\text{pos}}+\mathcal{L}_{\text{neg}}+\mathcal{L}_{\text{sim}} +  \alpha_{\text{aug}}\mathcal{L}_{\text{aug}}.
\end{equation}
$\alpha_{\text{aug}}$ is a hyper-parameters that controls the scale of $\mathcal{L}_{\text{aug}}$.

 \begin{table*}[t]
\centering
\resizebox{0.99\linewidth}{!}{
%
\begin{tabular}{c|l c|cccc cccc}
    \toprule[2pt]
    \multirow{2}{*}[-1.0ex]{\centering\arraybackslash\textbf{Setting}} & \multirow{2}{*}[-1.0ex]{\centering\arraybackslash\textbf{Model}}
    & \multirow{2}{*}[-1.0ex]{\centering\arraybackslash\textbf{Features}}
    & \multicolumn{8}{c}{\textbf{Captioning}} \\
\cmidrule(lr){4-11}& & 
    & \textbf{SODA\_c} & \textbf{METEOR} & \textbf{CIDEr} & \textbf{ROUGE-L} & \textbf{BLEU-1} & \textbf{BLEU-2} & \textbf{BLEU-3} & \textbf{BLEU-4} \\
\midrule

    
    \multirow{6}{*}{\centering\arraybackslash\textbf{Fully-Supervised}} 
    & DCE~\cite{ActivityNet} & C3D & -- & 5.69 & 12.43 & -- & 10.81 & 4.57 & 1.90 & 0.71 \\
    & DVC~\cite{DVC} & C3D & -- & 6.93 & 12.61 & -- & 12.22 & 5.72 & 2.27 & 0.73 \\
    & PDVC~\cite{PDVC} & C3D & 5.26 & 7.50 & 25.87 & -- & -- & -- & -- & 1.65 \\
    & Vid2Seq~\cite{vid2seq} & CLIP & 5.80 & 8.50 & 30.10 & -- & -- & -- & -- & -- \\
    & CM$^2$~\cite{cm2} & CLIP & 6.18 & 8.55 & 33.01 & -- & -- & -- & -- & 2.38 \\
    & E$^2$DVC~\cite{e2dvc} & CLIP & 6.13 & {8.57} & 33.63 & -- & -- & -- & -- & {2.43} \\
    \midrule 

    \multirow{6}{*}{\centering\arraybackslash\textbf{Weakly-Supervised}} 
    & WSDEC~\cite{WSDEC} & C3D & -- & 6.30 & 18.77 & 12.55 & 12.41 & 5.50 & 2.62 & 1.27 \\
    & ECG~\cite{ECG} & C3D & -- & 7.06 & 14.25 & -- & 11.85 & 5.64 & 2.71 & 1.33 \\
    & EC-SL~\cite{EC-SL} & C3D & -- & 7.49 & 21.21 & 13.02 & 13.36 & 5.96 & 2.78 & 1.33 \\
    & PWS-DVC$^*$~\cite{pws-dvc} & C3D & -- & 7.28 & 20.59 & 12.71 & -- & -- & -- & 1.35 \\
    & ILCACM~\cite{ge2025implicit} & CLIP & {6.08} & {8.48} & {33.42} & {14.77} & {15.36} & {8.12} & {4.17} & {2.26} \\
     \rowcolor{cGrey} & $\model$ (Ours) & CLIP & \textbf{6.29} & \textbf{8.63} & \textbf{35.38} & \textbf{15.29} & \textbf{15.86} & \textbf{8.30} & \textbf{4.27} & \textbf{2.30} \\
    \bottomrule[2pt]
\end{tabular}%
}
\caption{Comparison with state-of-the-art methods on ActivityNet validation set.
$*$ denotes results reported in~\cite{ge2025implicit}.
Our method outperforms the previous state-of-the-art weakly-supervised approach~\cite{ge2025implicit} and remarkably surpasses several fully-supervised methods~\cite{vid2seq,cm2,e2dvc}.
This demonstrates semantic alignment and caption augmentation can effectively replace temporal boundary supervision.
}
\label{table:main_caption} 
\end{table*}

\begin{table*}[t]
\centering
\label{tab:main}      
\resizebox{\textwidth}{!}{
%
\begin{tabular}{c|l c|ccccc|ccc}
    \toprule[2pt]
    \multirow{2}{*}[-1.0ex]{\centering\arraybackslash\textbf{Setting}} & \multirow{2}{*}[-1.0ex]{\centering\arraybackslash\textbf{Model}}
    & \multirow{2}{*}[-1.0ex]{\centering\arraybackslash\textbf{Features}}
    & \multicolumn{5}{c|}{\textbf{Captioning}} 
    & \multicolumn{3}{c}{\textbf{Localization}} \\
    \cmidrule(lr){4-8} \cmidrule(lr){9-11}
    & & 
    & \textbf{SODA\_c} & \textbf{METEOR} & \textbf{CIDEr} & \textbf{ROUGE-L} & \textbf{BLEU@N}
    & \textbf{R@AVG} & \textbf{P@AVG} & \textbf{F1} \\
    \midrule 
    
    \multirow{4}{*}{\centering\arraybackslash\textbf{Weakly-Supervised}} 
    & WSDEC$^*$~\cite{WSDEC} & C3D & 2.11 & 1.47 & 8.43 & -- & -- & -- & -- & -- \\
    & PWS-DVC$^*$~\cite{pws-dvc} & C3D & 3.14 & 2.48 & 9.81 & -- & -- & -- & -- & -- \\
    & ILCACM$^\dagger$~\cite{ge2025implicit} & CLIP & {3.60} & {3.41} & {13.49} & {4.75} & {2.59} & {17.76} & {18.01} & {17.88} \\
     \rowcolor{cGrey} & $\model$ (Ours) & CLIP & \textbf{4.08} & \textbf{3.63} & \textbf{14.61} & \textbf{5.42} & \textbf{2.94} & \textbf{20.76} & \textbf{21.13} & \textbf{20.94} \\
    \bottomrule[2pt]
\end{tabular}%
}
\caption{Comparison with other weakly-supervised methods on YouCook2 validation set. 
$\dagger$ means our re-implementation results, as the original paper does not report localization performance on this dataset.
Our approach also achieves the best captioning and localization performance among all weakly-supervised methods on this dataset.
BLEU@N denotes the average of BLEU-1 through BLEU-4.
}
\label{table:yc2}
\end{table*}

\begin{table}[h!]
\centering
\resizebox{\linewidth}{!}{
    \begin{tabular}{c|l c|ccc}
    \toprule[2pt]
    \multirow{2}{*}[-1.0ex]{\centering\arraybackslash\textbf{Setting}} 
    & \multirow{2}{*}[-1.0ex]{\centering\arraybackslash\textbf{Model}} 
    & \multirow{2}{*}[-1.0ex]{\centering\arraybackslash\textbf{Features}}
    & \multicolumn{3}{c}{\textbf{Localization}} \\
\cmidrule(lr){4-6}
    & & 
    & \textbf{R@Avg} & \textbf{P@Avg} & \textbf{F1} \\
    \midrule
    \multirow{5}{*}{\begin{tabular}[c]{@{}c@{}}\textbf{Fully-}\\ \textbf{Supervised}\end{tabular}} & SDVC~\cite{SDVC}        & C3D         & 55.58          & 57.57          & 56.56         \\
    & PDVC~\cite{PDVC}        & TSN         & 55.42          & 58.07          & {56.71}   \\
    & Vid2Seq~\cite{vid2seq}    & CLIP        & {52.70}    & {53.90}    & 53.29         \\ 
    & CM$^2$~\cite{cm2}       & CLIP        & {53.71}    & {56.81}    & 55.21 \\
    & E$^2$DVC~\cite{e2dvc}     & CLIP        & {54.67}    & {57.70}    & 56.14 \\
    
    \midrule
    \multirow{5}{*}{\begin{tabular}[c]{@{}c@{}}\textbf{Weakly-}\\ \textbf{Supervised}\end{tabular}} & WSDEC$^*$~\cite{WSDEC}    & C3D         & 29.57          & 59.33          & 39.18         \\
    & PWS-DVC$^*$~\cite{pws-dvc}  & C3D         & 40.85          & 55.82          & 47.09         \\
    & ILCACM~\cite{ge2025implicit}    & C3D         & 53.09          & {59.20} & 55.98         \\
    & ILCACM~\cite{ge2025implicit}    & CLIP        & {53.72} & 58.92          & {56.20} \\
     \rowcolor{cGrey} & $\model$ (Ours)       & CLIP        & \textbf{54.39} & \textbf{59.87}   & \textbf{57.00} \\ \bottomrule[2pt]
    \end{tabular}
}
\caption{Comparison of the event localization performance across various fully-supervised and weakly-supervised models on ActivityNet validation set. 
$\model$ achieves state-of-the-art
performance in most of the localization metrics.
}
\label{table:main_localize}
\end{table}
\section{Experiments}
\subsection{Experimental Settings.}
\label{subsec: setting}
\noindent\textbf{Datasets.}
Our experiments are conducted on two standard benchmarks for DVC. 
\textbf{ActivityNet Captions}~\cite{ActivityNet} dataset is a larger collection of 20k untrimmed instructional videos, which average 120 seconds in length and are annotated with 3.7 localized events per video.
\textbf{YouCook2}~\cite{youcook2} dataset comprises approximately 2K untrimmed culinary videos, each with an average duration of 320 seconds and annotated with 7.7 localized language sentences. 

\noindent\textbf{Evaluation Metrics.}
We evaluate our model's performance on the two primary sub-tasks of DVC: caption generation and event localization. To evaluate caption quality, we adopt the official evaluation tool~\cite{ActivityNet} to report standard metrics: METEOR~\cite{banerjee2005meteor}, CIDEr~\cite{vedantam2015cider}, ROUGE-L~\cite{lin2004rouge} and BLEU-N~\cite{papineni2002bleu}. 
To assess the narrative quality of the generated descriptions, we also employ the SODA\_c metric~\cite{fujita2020soda}. 
For the event localization task, we measure the mean Average Precision (mAP), mean Average Recall (mAR), and the F1 score. 
In line with the standard protocol, all metrics for both tasks are averaged over a set of Intersection over Union (IoU) thresholds: \{0.3, 0.5, 0.7, 0.9\}.

\noindent\textbf{Implementation Details.}
Following previous work~\cite{ge2025implicit}, we use the Distilled-GPT2 model~\cite{GPT2} as our caption decoder and the AdamW optimizer for training. For the ActivityNet Captions dataset~\cite{ActivityNet}, we set the initial learning rate to 1e-4 and train for 10 epochs for both the captioning and localization stages. For the YouCook2 dataset~\cite{youcook2}, we use an initial learning rate of 5e-5 and train for 5 and 15 epochs for the captioning and localization stages, respectively. 
For the synthetic caption generation, we utilize the Qwen3-8B~\cite{yang2025qwen3}.
The hyperparameters are fixed across all experiments with  $\Delta=0.1$, $w^{inter}=0.6$ and $\alpha_{\text{aug}}=0.25$.
\subsection{Comparison with State-of-the-Art.}
The results for the captioning and localization tasks on the ActivityNet Captions dataset are presented in Table~\ref{table:main_caption} and Table~\ref{table:main_localize}, respectively. 
Our proposed model, $\model$, demonstrates the highest overall performance in both tasks. 
As shown in Table~\ref{table:main_caption}, our model achieves a CIDEr of 35.38, surpassing the previous state-of-the-art WSDVC model~\cite{ge2025implicit} and consistently outperforming it across all other captioning metrics. 
In the localization task, $\model$ also shows the best performance, achieving the best F1 score of 57.00, which is driven by the highest recall (54.39) and precision (59.87) among all weakly-supervised methods.
Remarkably, our weakly-supervised approach outperforms the fully-supervised method~\cite{cm2,e2dvc} across the majority of metrics, despite not using temporal boundary annotations during training.
These results validate that our semantically-aware masks successfully emphasize event-relevant regions, enabling the generation of high-quality captions that accurately describe target events.
In Table~\ref{table:yc2}, $\model$ also achieves the highest captioning and localization scores on YouCook2~\cite{youcook2} compared to other WSDVC methods.
These consistent improvements across datasets demonstrate our model's strong performance in both tasks.
\begin{figure}[t]
\centering\includegraphics[width=\linewidth]{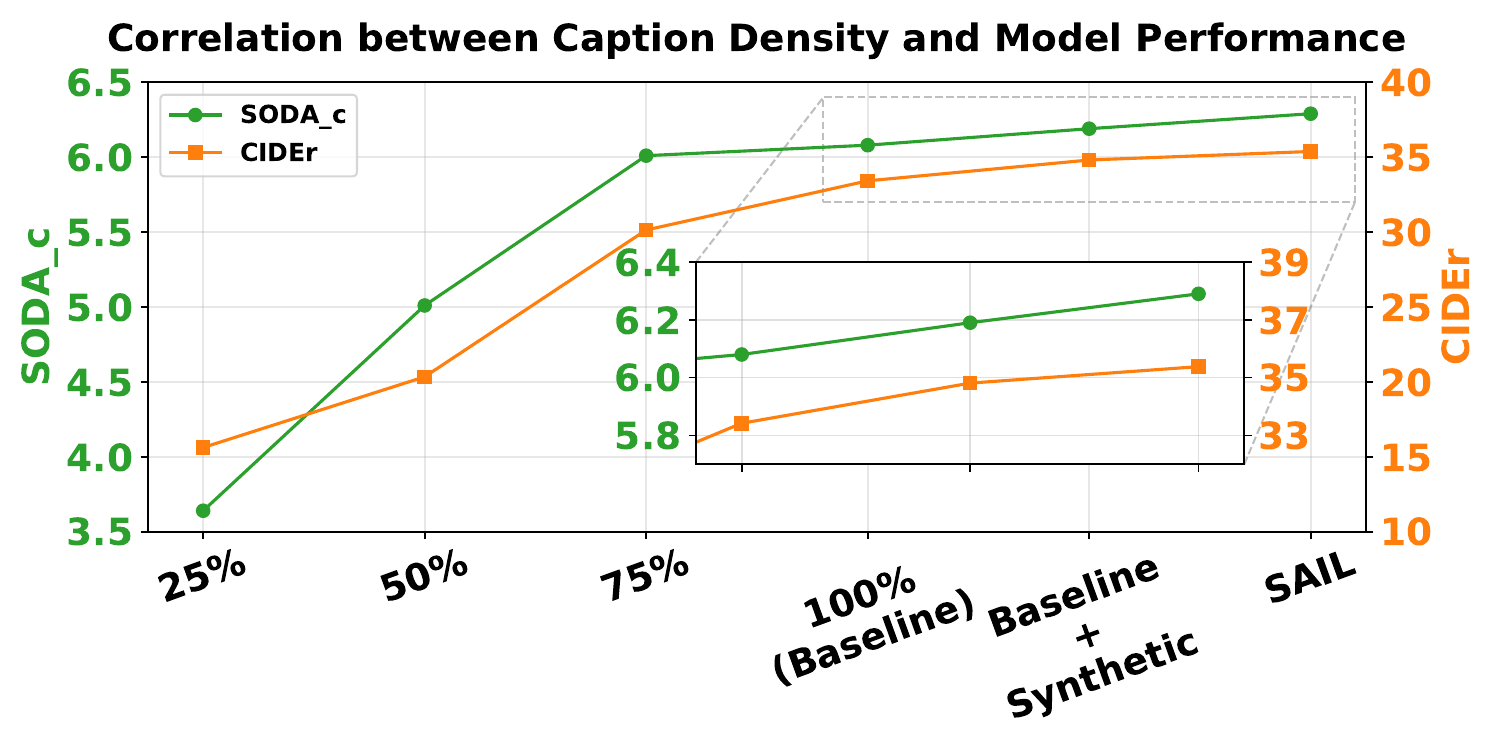}
\vspace{-5mm}
\caption{Impact of caption density on model performance.
Performance decreases consistently as annotation density reduces from 100\% to 25\% (left), highlighting annotation sparsity as a critical challenge.
Densifying supervision through LLM-generated synthetic captions (right) improves performance.}
\label{fig:dataset_analysis}
\end{figure}
\begin{figure*}[t]
\centering\includegraphics[width=\linewidth]{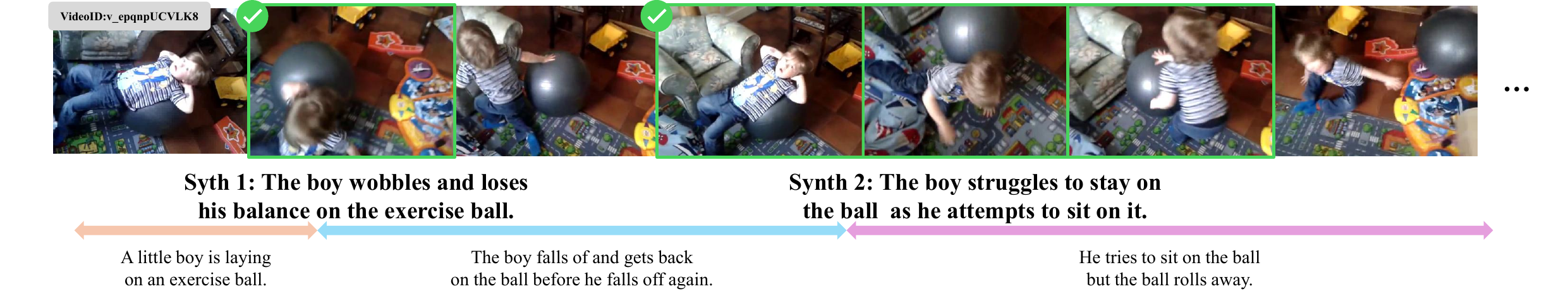}
\caption{Synthetic caption's qualitative results. Our synthetic captions effectively capture potential intermediate events occurring between consecutive ground-truth annotations.}
\label{fig:LLMqual}
\end{figure*}

\noindent\textbf{Correlation between Caption Density and Model Performance.}
To analyze the impact of event annotation density on WSDVC performance, we conduct experiments on the baseline method~\cite{ge2025implicit} using different proportions of ground-truth captions: 25\%, 50\%, 75\%, and 100\% (Baseline~\cite{ge2025implicit}).
As shown in~\Cref{fig:dataset_analysis}, the model's performance drops consistently as the number of annotated events decreases.
Notably, as WSDVC must learn event localization using only captions without temporal boundaries, sparse captions significantly degrade both the captioning quality (CIDEr) and the inter-sentence temporal alignment (SODA\_c).
The right side shows the performance when event annotations are densified through our LLM-based augmentation strategy.
Beyond the original ground-truth descriptions, our synthetic caption generation improves performance even when simply incorporating them into the baseline method.

\subsection{Ablation Studies.}
\noindent\textbf{Component Ablation.}
In Table~\ref{table:component_ablation}, we analyze the contribution of each component. We observe that applying a similarity-aware mask alone improves CIDEr by {+1.76} and SODA\_c by {+0.21}, indicating that similarity-based loss effectively produces masked features that are well-aligned with caption content. 
Similarly, employing synthetic captions with inter-event masks provides improvements for both captioning and localization, demonstrating the effectiveness of dense event supervision. 
Combining both components achieves the best performance, improving all metrics. This synergy enables our model to learn more discriminative event representations.

\noindent\textbf{Data Scale Ablations.}
As shown in Table~\ref{table:data_scale}, incorporating even a small fraction (25\%) of the synthetic captions leads to a performance gain over the baseline in both captioning and localization.
Also, we observe that the performance in both tasks improves monotonically as the ratio of synthetic captions increases. 
Our full model (100\%) achieves the highest scores across all metrics.
Regardless of the ratio used, every configuration of our method consistently and substantially outperforms the baseline.

\begin{table}[t]
    \centering
    \resizebox{0.43\textwidth}{!}{
        \begin{tabular}{c c|c c c|c}
        \toprule[2pt]
        
        \multirow{2}{*}[-1.0ex]{\textbf{Sim}}
        & \multirow{2}{*}[-1.0ex]{\textbf{Synth}}
        & \multicolumn{3}{c|}{\textbf{Captioning}}
        & \multicolumn{1}{c}{\textbf{Localization}} \\
        \cmidrule(lr){3-5} \cmidrule(lr){6-6}
        & 
        & \textbf{S\_c}
        & \textbf{M}
        & \textbf{C}
        & \textbf{F1} \\
        \midrule
            \ding{55} & \ding{55} & 6.08 & 8.48 & 33.42 & 56.20 \\
            
            \ding{51} & \ding{55} &6.27  & {8.58} & 35.18 & 56.89 \\

            \ding{55} & \ding{51} & \textbf{6.29} &8.52  & 34.92 & 56.79 \\
            
            \rowcolor{cGrey}\ding{51} & \ding{51} & \textbf{6.29} & \textbf{8.63} & \textbf{35.38} & \textbf{57.00} \\
            \bottomrule[2pt]
        \end{tabular}
    }
    \caption{Ablation study on our key components on ActivityNet. \textbf{Sim} denotes similarity-aware mask, and \textbf{Synth} denotes Synthetic Captions with inter mask.
    S\_c, M, and C denote SODA\_c, METEOR, and CIDEr, respectively.}
    \label{table:component_ablation}
\end{table}
\begin{table}[t]
    \centering
    \resizebox{0.45\textwidth}{!}{
    \begin{tabular}{l|ccc|c} 
        \toprule[2pt]
        \multirow{2}{*}[-0.5ex]{\textbf{Method}} & \multicolumn{3}{c|}{\textbf{Captioning}} & \multicolumn{1}{c}{\textbf{Localization}} \\
        \cmidrule(lr){2-4} \cmidrule(lr){5-5} 
        & \textbf{S\_c} & \textbf{C} & \textbf{R} & \textbf{F1} \\
        \midrule
        0\% (Baseline) & 6.08 & 33.42 & 14.77 & 56.20 \\
        
        25\%          & 6.08 & 33.76 & 15.02 & 56.38 \\
        
        50\%          & \underline{6.27} & 34.19 & 15.00 & 56.39 \\
        
        75\%          &6.23  & \underline{34.66} & \underline{15.09} & \underline{56.91} \\
        
        \rowcolor{cGrey} 
        100\% (Ours)   & \textbf{6.29} & \textbf{35.38} & \textbf{15.29} & \textbf{57.00} \\
        \bottomrule[2pt]
    \end{tabular}
    }
    \caption{Ablation study on the amount of synthetic captions used. The results suggest that the benefits of synthetic captions are evident even at a small scale.
    R denotes ROUGE-L.}
    \label{table:data_scale}
\end{table}


            
\begin{table}[t]
    \centering
    \resizebox{0.45\textwidth}{!}{ 
        \begin{tabular}{c|ccccc}
            \toprule[2pt]
            \multirow{2}{*}[-1.0ex]{\textbf{Method}} & \multicolumn{5}{c}{\textbf{Captioning}} \\
            \cmidrule(lr){2-6}
            & \textbf{S\_c} & \textbf{M} & \textbf{C} & \textbf {R} & \textbf {B@4}\\
            \midrule

            ILCACM~\cite{ge2025implicit} & {6.08}   & \underline{8.48} & {33.42} & 14.77 & \underline{2.26}\\

            Synth + HN &\underline{6.24}  &8.44 & \underline{33.84} &\underline{15.10}  & {2.02} \\
            
            Synth + Inter & \textbf{6.29} & \textbf{8.63}  & \textbf{35.38}  & \textbf{15.29} & \textbf{2.30}\\
            
            \bottomrule[2pt]
        \end{tabular}
    }
    \caption{Ablation study on synthetic caption utilization strategies.
Leveraging synthetic captions indirectly via an inter-mask mechanism yields the best overall performance.
B$@$4 denotes BLEU-4.
    }
    \label{table:synth_cap_utilize} 
\end{table}
\begin{table*}[t]
\centering

\resizebox{0.95\linewidth}{!}{
\begin{tabular}{c|c|cccc|ccc} 
\toprule[2pt]
\multirow{2}{*}[-1.0ex]{\centering\arraybackslash\textbf{Mask Type}}
& \multirow{2}{*}[-1.0ex]{\centering\arraybackslash\textbf{Model}}
& \multicolumn{4}{c|}{\textbf{Captioning}}
& \multicolumn{3}{c}{\textbf{Localization}} \\
\cmidrule(lr){3-6} \cmidrule(lr){7-9}
& & \textbf{SODA\_c} & \textbf{METEOR} 
& \textbf{CIDEr} & \textbf{ROUGE-L}& \textbf{R@Avg} & \textbf{P@Avg} & \textbf{F1} \\
\midrule
\multirow{2}{*}{Gaussian Mask} 
& ILCACM~\cite{ge2025implicit} & 6.08 & 8.48 & 33.42 & 14.77 & 53.72 & 58.92 & 56.20 \\
& \model~(Ours) & \textbf{6.29} & \textbf{8.63} & \textbf{35.38} 
& \textbf{15.29} & \textbf{54.39} & \textbf{59.87} & \textbf{57.00} \\
\midrule

\multirow{2}{*}{Hard Binary Mask} 
& ILCACM$^{\dagger}$~\cite{ge2025implicit} & 3.89 & 6.52 & 16.96 & \textbf{11.24} & 35.76 & 50.85 & 41.91 \\
& \model~(Ours) & \textbf{4.09} & \textbf{6.53} & \textbf{17.54} & {11.07} & \textbf{35.81} & \textbf{51.00} & \textbf{42.00} \\
\midrule


\multirow{2}{*}{Cauchy Mask} 
& ILCACM$^{\dagger}$~\cite{ge2025implicit} & 6.02 & 8.32 & 32.22 & 14.50 & 53.06 & 58.87 & 55.81 \\
& \model~(Ours) & \textbf{6.18} & \textbf{8.66} & \textbf{33.85} & \textbf{15.06} & \textbf{54.36} & \textbf{59.11} & \textbf{56.63} \\
\bottomrule[2pt]
\end{tabular}
}
\caption{Ablation study of various mask designs. \model~demonstrates consistent performance improvements over the baseline across all mask designs. $\dagger$ means our re-implementation localization results, as the original paper does not report localization performance.}
\label{table:mask_type}
\end{table*}
\noindent\textbf{{Synthetic Caption Utilization.}}
We conduct experiments on different strategies for utilizing synthetic captions. 
First, we include synthetic captions and their corresponding inter-event masked video features when computing hard negatives for positive pairs in $\mathcal{L}_{\text{sim}}$. 
Second, as described in our method, we treat synthetic captions and inter-event masked features as auxiliary signals, computing $\mathcal{L}_{\text{sim}}$ only on the ground-truth annotation set while using synthetic captions as supplementary supervision through $\mathcal{L}_{\text{aug}}$.
As shown in Table~\ref{table:synth_cap_utilize}, simply incorporating synthetic captions (Synth + HN) already outperforms ILCACM~\cite{ge2025implicit}, demonstrating the effectiveness of densifying supervision signals.
Moreover, our auxiliary guidance approach, which provides a separate alignment signal, achieves the best results, validating our design choice of treating synthetic captions as soft guidance rather than hard constraints.

\noindent\textbf{Mask-type Ablations.}
In Table~\ref{table:mask_type}, we evaluate different mask types used for video feature masking. 
Gaussian masks achieve the best performance at 35.38 CIDEr and 6.29 SODA\_c, substantially outperforming Hard binary and Cauchy alternatives. 
Notably, all mask types show consistent improvements when combined with our proposed framework compared to the baseline~\cite{ge2025implicit}, demonstrating the effectiveness of our similarity-aware guidance and the use of synthetic captions with inter-event masks.

\begin{figure}[t]
\centering\includegraphics[width=1.0\columnwidth]{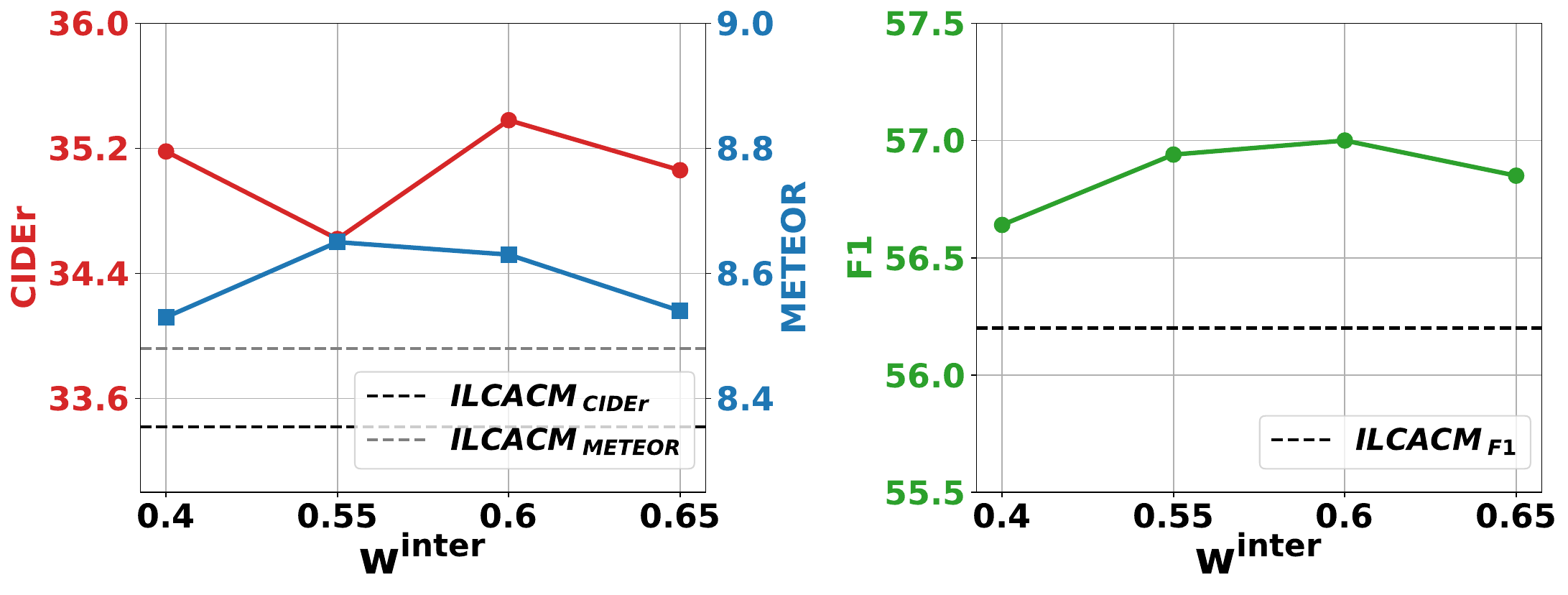}
\caption{Impact of hyper-parameter $w^{inter}$ for inter masks on captioning and localization performance.}
\vspace{-1mm}
\label{fig:hyperparam1}
\end{figure}
\begin{table}[t]
    \centering
    \resizebox{0.48\textwidth}{!}{ 
        \begin{tabular}{l|ccc}
            \toprule[2pt]
            \textbf{Method} & \textbf{Train time} &
            \textbf{Inference time}&
            \textbf{GPU usage} 
            
            \\
            \midrule

            ILCACM & {1H 38M 27S} & {7M 11S} &{33.06 \text{GiB}}   
            \\
            
            $\model$ (Ours)  &{1H 41M 59S} & {7M 01S} &{33.11 \text{GiB}} 
            \\

            \bottomrule[2pt]
        \end{tabular}
    }
    \caption{Computational cost comparison against the baseline~\cite{ge2025implicit}.}
    \label{table:rebuttal_cost}
\end{table}
\noindent\textbf{{Performance-computation trade-off.}}
We analyze the computational cost of our method in comparison to the baseline. 
All metrics are averaged over five independent runs.
As shown in Table~\ref{table:rebuttal_cost}, the training time, inference time, and GPU memory consumption of $\model$ remain nearly identical to those of the baseline~\cite{ge2025implicit}.
Our approach only introduces a lightweight dot-product operation, incurring negligible overhead. 
Furthermore, the LLM-based caption augmentation is highly efficient; it is a one-time preprocessing step that operates solely on textual data, requiring only 0.28 hours to complete for the ActivityNet dataset.
\begin{figure}[h]
\centering\includegraphics[width=0.99\columnwidth]{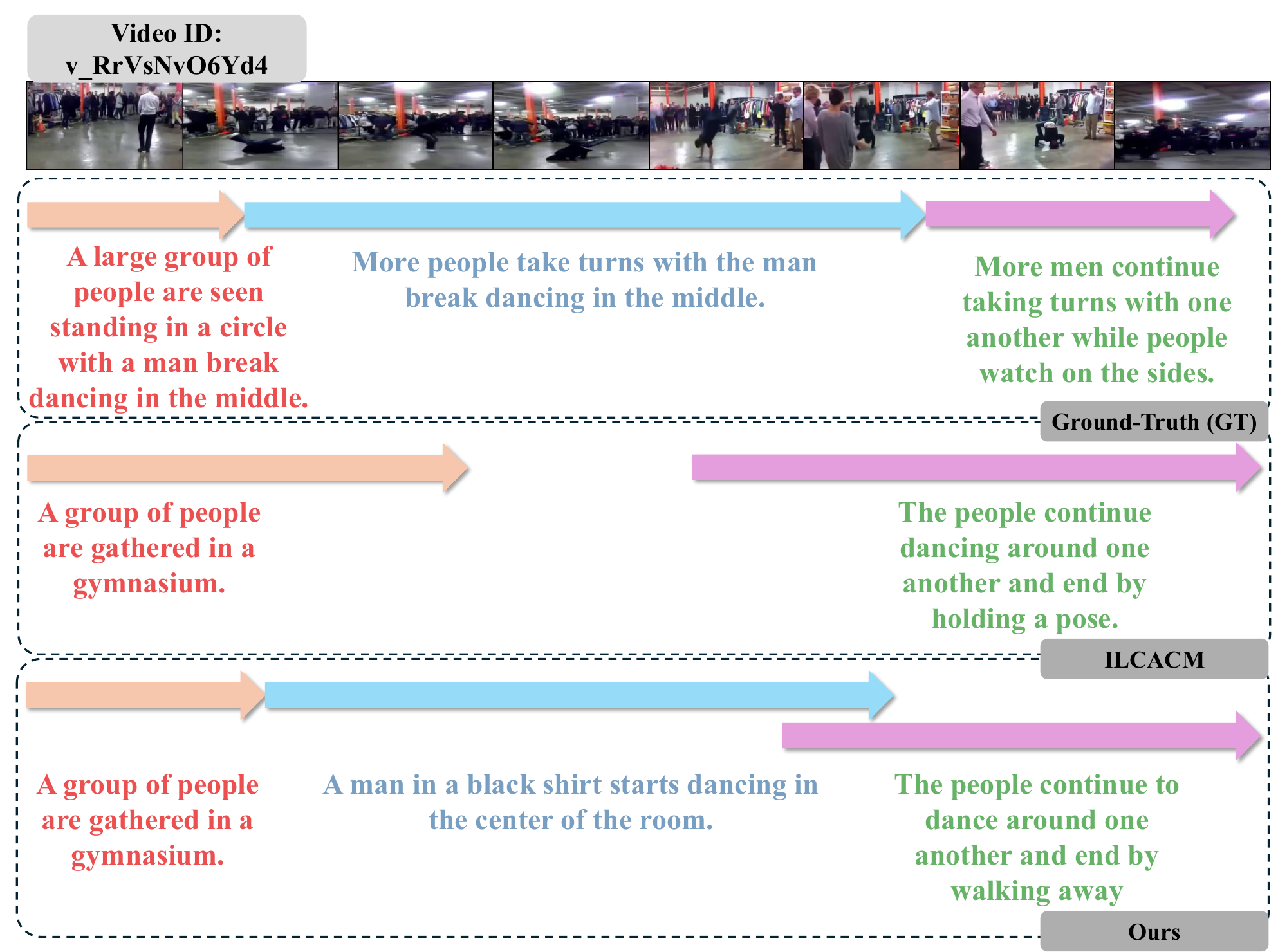}
\caption{A qualitative result from the ActivityNet validation set.}
\label{fig:qualitative}
\end{figure}

\noindent\textbf{Hyper-parameter Ablations.}
We conduct ablation experiments to evaluate captioning and localization scores under different values of $w^{\text{inter}}$, which controls the inter-event mask width, and maintain robust performance across different values(\Cref{fig:hyperparam1}).
We choose $w^{\text{inter}}=0.6$ for its balanced captioning and localization scores. 
Additional hyperparameter experiments are in the supplementary material.

\noindent\textbf{Qualitative Results.}
~\Cref{fig:LLMqual} shows that our synthetic captions accurately describe the events situated between ground-truth annotations.
Additionally,~\Cref{fig:qualitative} shows that our $\model$ predicts event boundaries and captions that align well with the video content. 
\label{sec:exp}

 \section{Conclusion}
\label{sec:con}
{In this work, we propose $\model$, a novel weakly supervised approach to dense video captioning. 
Previous methods apply Gaussian masks to merely different temporal segments without accounting for semantic alignment between events and visual regions, leading to sub-optimal results. 
Our model addresses this limitation by leveraging cross-modal similarity to guide masks toward event-relevant regions. Crucially, we augment sparse GT annotations with LLM-augmented events, enabling our model to learn from a richer set of event-region alignments.
Extensive experiments on ActivityNet and YouCook2 demonstrate the effectiveness of our proposed method.
In future work, we plan to investigate methods for mitigating data scarcity across a diverse range of multimodal tasks~\cite{sida,sync,catchphrase,ifcap}, beyond the DVC task.}

\noindent\textbf{Acknowledgments.} 
This was partly supported by the Institute of Information \& Communications Technology Planning \& Evaluation (IITP) grant funded by the Korean government(MSIT) (No.RS-2020-II201373, Artificial Intelligence Graduate School Program(Hanyang University)) and the Institute of Information \& Communications Technology Planning \& Evaluation (IITP) grant funded by the Korean government(MSIT) (No.RS-2025-02215122, Development and Demonstration of Lightweight AI Model for Smart Homes).
 {  \small
    \bibliographystyle{ieeenat_fullname}
    \bibliography{main}}

\clearpage
\setcounter{page}{1}
\maketitlesupplementary
\appendix

\section{Implementation Details}
\label{sec:implement}
\noindent\textbf{Backbone.}
We use CLIP ViT-L/14~\cite{clip} to extract visual features from video frames.
For ActivityNet Captions, we uniformly sample 32 frames per video, while for YouCook2, we sample 100 frames.
Text features are extracted using the CLIP text encoder.

\noindent\textbf{LLM Settings.}
We employ the Qwen3-8B model~\cite{yang2025qwen3} for synthetic caption generation with the following hyperparameters: maximum new tokens = 50, temperature = 0.7.

\noindent\textbf{Event Queries.}
The number of event queries is set to 14 for ActivityNet Captions and 18 for YouCook2, respectively, to accommodate the different annotation densities of the two datasets.

\noindent\textbf{Mask Generation Module.}
Following~\cite{ge2025implicit}, our mask generation module consists of a transformer-based decoder that processes learnable event queries.
Specifically, the transformer layers take the event queries as input and produce event-aware embeddings, which are then passed through separate fully-connected layers to predict the temporal center $c_i$ and width $w_i$ for each event.
These values are used to construct Gaussian masks via Eq.~\eqref{eq:mask}.

\section{Inference Details}
\label{sec:inference}
At test time, our model employs the same inference procedure as~\cite{ge2025implicit} to generate temporally localized event captions.
First, we obtain an initial set of event descriptions by feeding the complete video embedding into the caption decoder with a global context prompt (“[FULL]").
The model adaptively infers the number of events present in the video during this phase.
Subsequently, we apply the mask generation module to predict temporal parameters (centers and widths) for each identified event, from which we derive Gaussian attention masks.
Finally, we enhance caption quality through a refinement mechanism: each event's masked video representation is decoded using event-specific conditioning (“[MASK] 1 events:") to produce more precise and contextually appropriate descriptions.

\begin{figure}[t]
\centering\includegraphics[width=1.0\columnwidth]{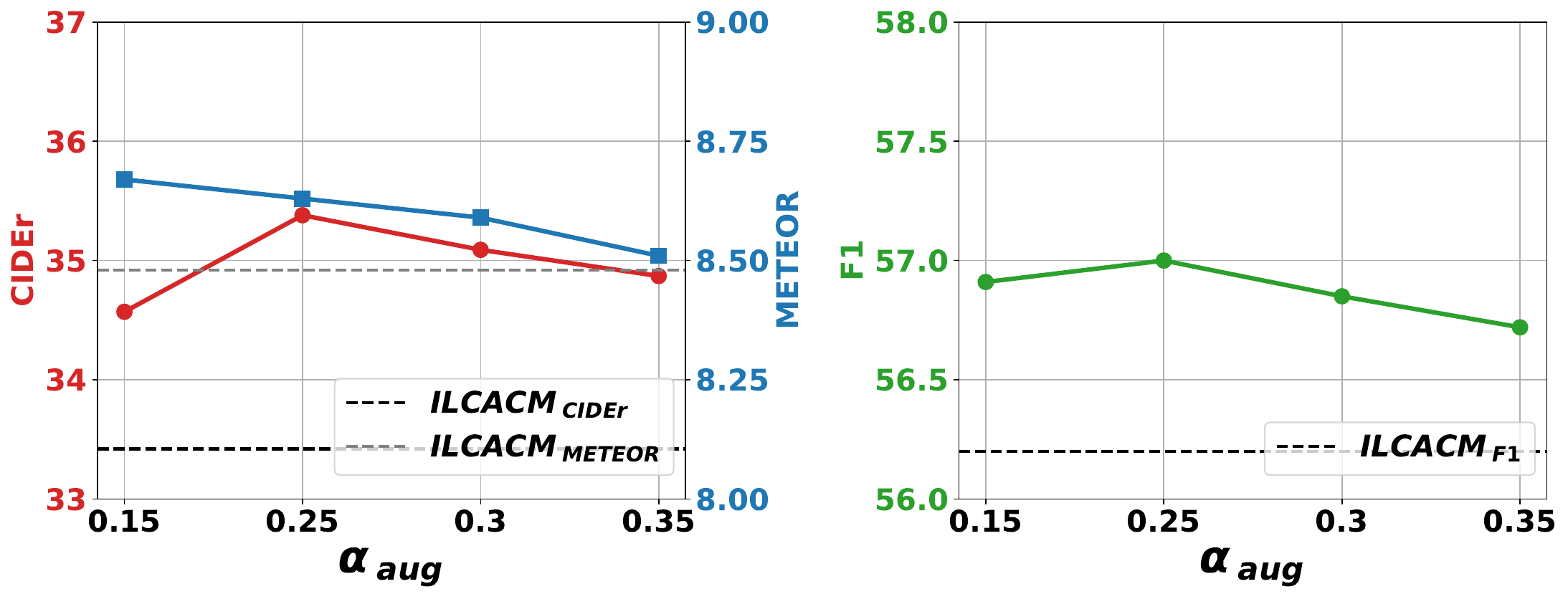}
\caption{Impact of hyper-parameter $\alpha_\text{aug}$ for scaling $\mathcal{L}_{\text{aug}}$ on captioning and localization performance.}
\vspace{-1mm}
\label{fig:hyperparam3}
\end{figure}
\subsection{Hyper-parameter Ablations.}
We conduct additional hyper-parameter ablation experiments to evaluate captioning and localization performance under different values of $\alpha_{\text{aug}}$.
~\Cref{fig:hyperparam3} shows the captioning and localization scores under different values of $\alpha_{\text{aug}}$.
The hyperparameter $\alpha_{\text{aug}}$ serves to scale the auxiliary loss to a comparable magnitude with the main losses.
The results demonstrate robust performance across different values, consistently outperforming the baseline~\cite{ge2025implicit} regardless of the specific weight choice.
We select $\alpha_{\text{aug}}$ as 0.25.

\begin{figure*}[h]
\centering\includegraphics[width=2\columnwidth]{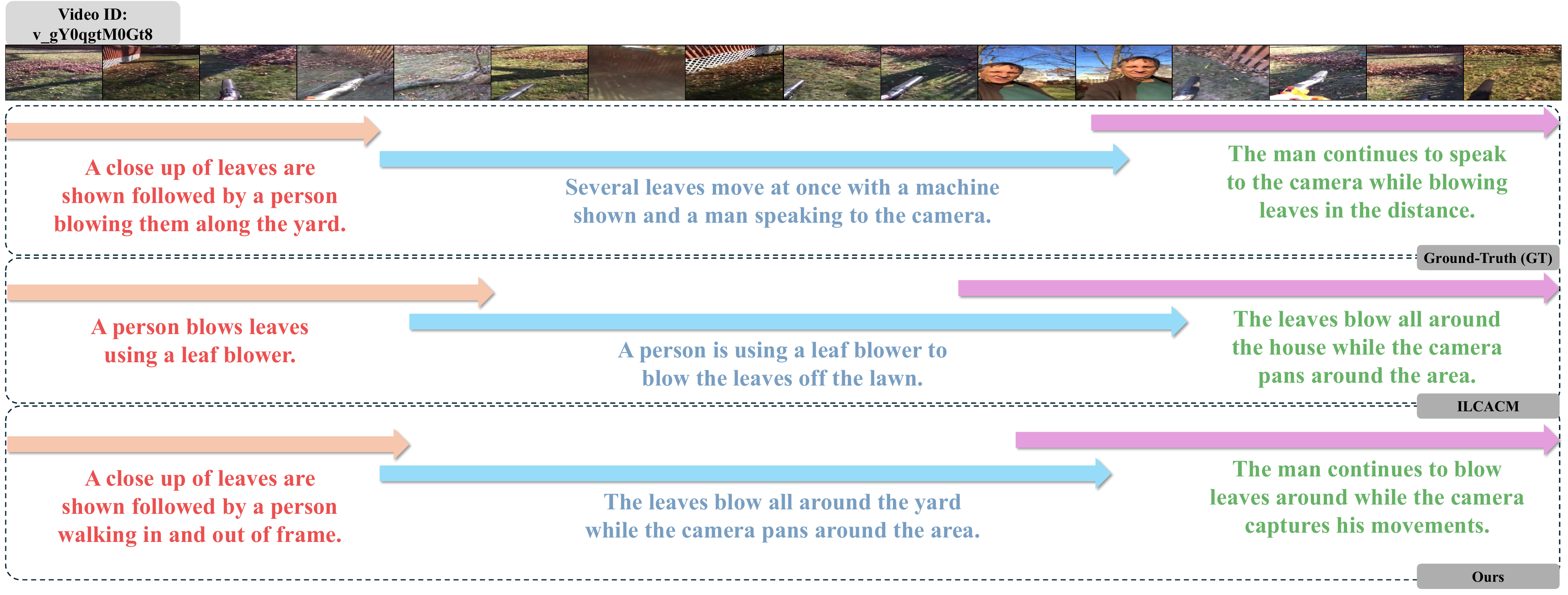}
\caption{Another qualitative result from the ActivityNet validation set.}
\label{fig:qualitative2}
\end{figure*}

\subsection{Qualitative Results.}
In this section, we provide additional qualitative results to demonstrate the effectiveness of our approach.
As shown in~\Cref{fig:qualitative2}, our method generates captions that are more closely aligned with ground-truth descriptions and achieves more accurate event localization compared to the baseline method~\cite{ge2025implicit}.

\subsection{Augmented Captions Quality.}
\begin{figure*}[t]
\centering\includegraphics[width=2.0\columnwidth]{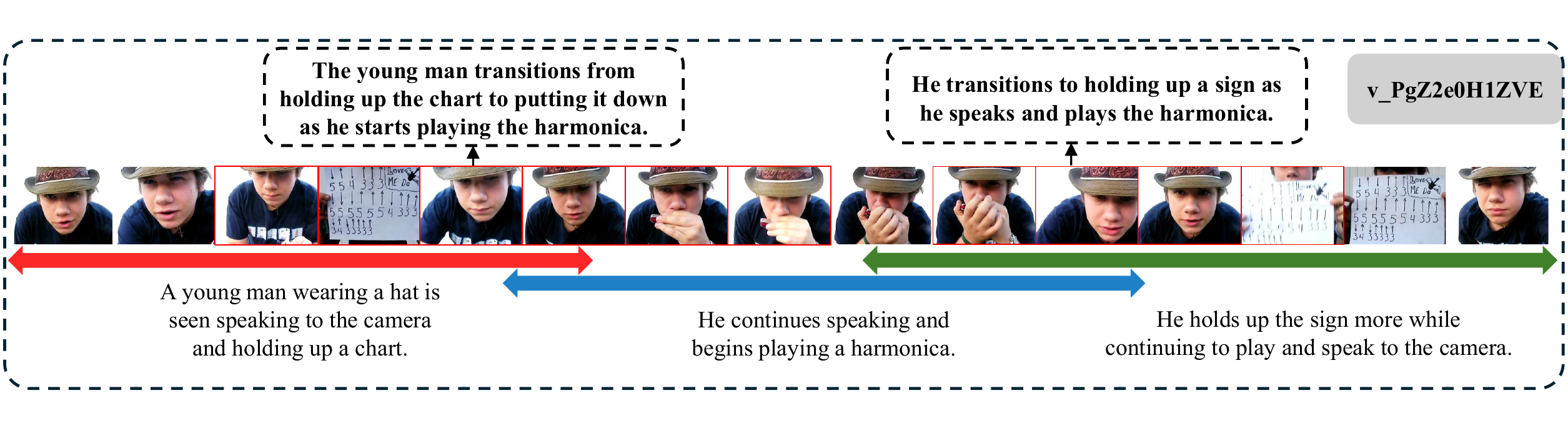}
\caption{Qualitative results LLM augmentation captions. Our LLM-generated synthetic captions effectively describe the intermediate events occurring between consecutive ground-truth annotations.
}
\vspace{-1mm}
\label{fig:LLM_qual3}
\end{figure*}
We present additional qualitative results for LLM-generated captions in~\Cref{fig:LLM_qual3}.
As shown, our LLM-generated captions effectively describe potential intermediate events such as “holding up a chart" and “playing the harmonica" that bridge consecutive ground-truth annotations.
The generated captions are contextually coherent and semantically appropriate, demonstrating the LLM's capability to infer plausible transitional events from textual context.
\subsection{Mask Quality.}
As discussed earlier, the baseline method~\cite{ge2025implicit} focuses solely on ensuring that masks cover different temporal regions without considering semantic alignment with their corresponding events.
If this is indeed the case, masks should exhibit similar width values to simply partition the video into distinct segments.
\begin{table}[t]
    \centering
    \resizebox{0.4\textwidth}{!}{ 
        \begin{tabular}{c|ccc}
            \toprule[2pt]
            \textbf{Method} & \textbf{Mean} & \textbf{Min} & \textbf{Max} \\
            \midrule
            ILCACM~\cite{ge2025implicit} & 0.3489 & 0.2690 & {0.3900} \\
            
            $\model$ & \textbf{0.3535} & \textbf{0.2549} & \textbf{0.3914} \\
            
            \bottomrule[2pt]
        \end{tabular}
    }
    \caption{
Analysis of mask width diversity during training.
We report the mean (Mean), minimum (Min), and maximum (Max) standard deviation of mask widths across all training videos.
    }
    \label{table:mask_std}
\end{table}
To investigate this hypothesis, we calculate the standard deviation of mask widths across all videos in the training set and report the mean, minimum, and maximum standard deviation values in~\Cref{table:mask_std}.
As shown in the results, our similarity-aware mask construction generates masks with varying widths that accurately capture event-specific temporal characteristics.

\begin{figure*}[t]
\centering\includegraphics[width=2.0\columnwidth]{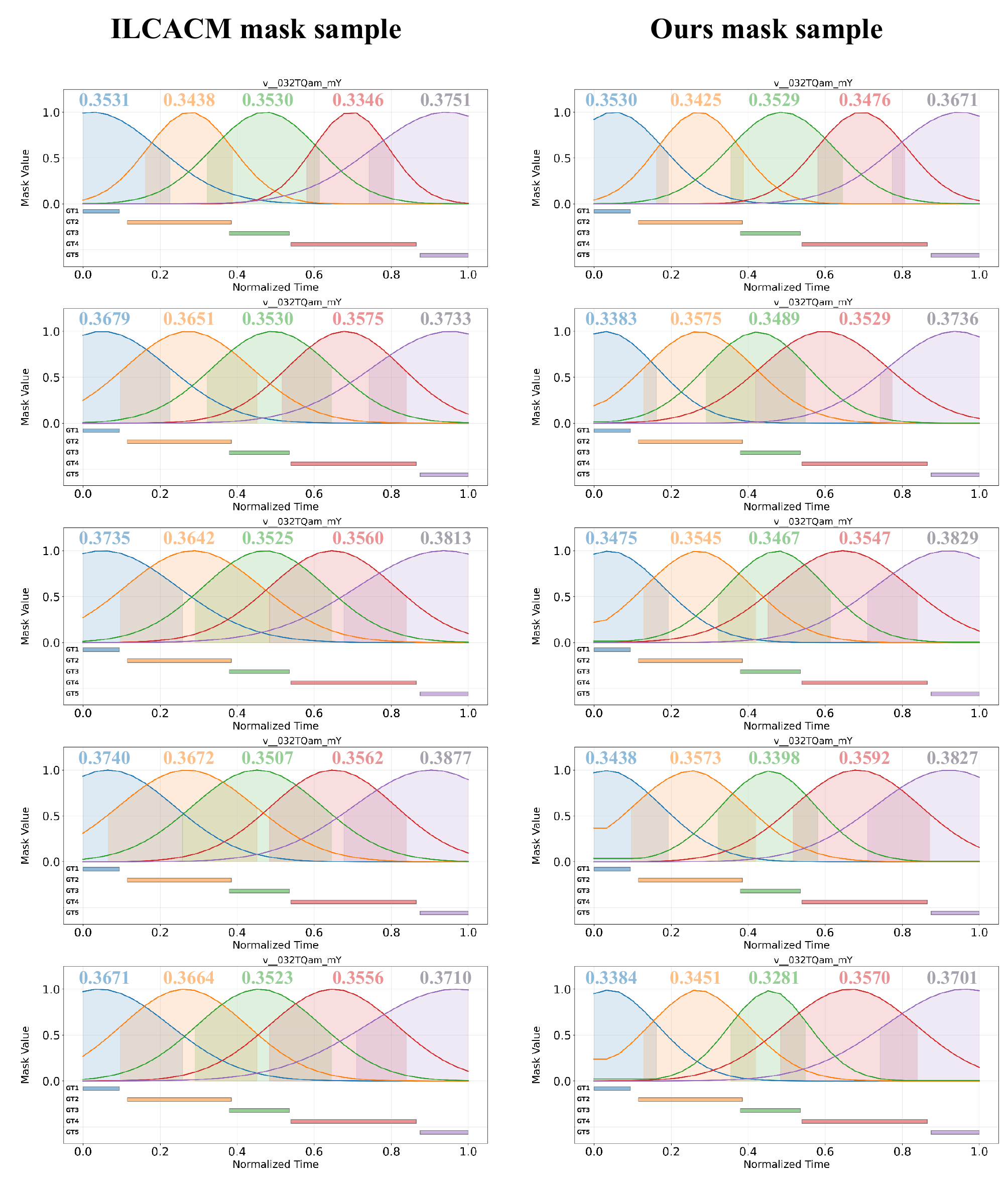}
\caption{We visualize the mask outputs from the mask generation module during training and annotate each mask with its corresponding width value. ILCACM maintains almost constant mask widths throughout training, whereas $\model$ shows progressive adaptation: masks for shorter events (blue, green) decrease in width, while masks for longer events (red) increase, aligning with actual event durations.
}
\vspace{-1mm}
\label{fig:mask_qual}
\end{figure*}

Additionally, qualitative results in Figure~\ref{fig:mask_qual} show that our approach produces event boundaries that align more closely with the non-uniform ground-truth timestamps compared to the baseline's uniform partitioning.
The baseline method focuses primarily on distributing masks evenly across the video, resulting in masks with uniform widths regardless of actual event durations.
For instance, the mask width for event 3 in ILCACM remains nearly constant throughout training epochs (0.3530 → 0.3530 → 0.3525 → 0.3507 → 0.3523), showing minimal adaptation to the event's actual temporal extent.
In contrast, our method learns masks by considering semantic alignment with corresponding event captions, enabling adaptive width adjustment.
As shown in~\Cref{fig:mask_qual}, the mask width for event 3 in our method progressively decreases during training (0.3529 → 0.3489 → 0.3467 → 0.3398 → 0.3281), demonstrating that it adapts to align with the event's shorter duration.
These results validate that our similarity-aware guidance enables masks to learn event-specific temporal characteristics rather than converging to uniform partitioning, leading to more accurate event localization aligned with ground-truth boundaries.
\section{Prompt Details}
\label{sec:prompt}
In~\Cref{algo:prompt}, we provide our instructions to the LLM for creating scene descriptions between consecutive events.
We specify five instructions: (1) analyzing both preceding and succeeding captions, (2) inferring the most probable transitional action or change of state, (3) maintaining consistency with the original annotations, (4) ensuring the generated event is highly plausible, and (5) adhering to the specified output format.
This structured approach helps the LLM concentrate on the specific temporal gap between events.

\begin{algorithm*}[t]
\caption{Our instruction prompts for LLM}

\SetAlgoLined
\DontPrintSemicolon
\textbf{SYSTEM PROMPT} 

You are a “Video Context Inference Expert," an AI specialized in analyzing sequences of video event captions. Your primary goal is to generate one new, plausible caption for the event that likely occurred *between* the two provided captions, creating a smooth and logical narrative flow.

1.  **Context is Key**: Deeply analyze the preceding and succeeding captions. The generated caption must serve as a logical bridge, connecting the two events seamlessly.

2.  **Infer the Unseen Action**: Do not simply rephrase or combine the given captions. Your task is to infer the most probable *transitional action* or *change of state*. Focus on the single most important action that connects the two moments.

3.  **Maintain Consistency**: The style, tone, and level of detail of your generated captions should match the input captions. They should be concise, descriptive, and written from the perspective of an objective observer.

4.  **Plausibility over Creativity**: The generated event must be highly plausible. Avoid introducing new elements that cannot be reasonably inferred.

5.  **Output**: Provide ONLY the single generated caption text, without any labels or explanations.
\BlankLine
\textbf{USER PROMPT} 
Analyze the following two consecutive video captions and generate a single, concise caption that describes the most plausible event happening between them.

**Caption 1:**:\{caption1\}

**Caption 2:** :\{caption2\}

Based on the rules, what is the single event that connects these two captions?
\label{algo:prompt}
\end{algorithm*}

\end{document}